\pdfoutput=1
\documentclass[journal]{IEEEtran}
\usepackage{amsmath}
\usepackage{amsthm}
\DeclareMathOperator*{\argmax}{argmax}
\newtheorem{theorem}{Theorem}

\usepackage{cases}
\usepackage{graphicx}
\usepackage{hyperref}
\usepackage{color,soul}
\usepackage{booktabs} 
\usepackage{subcaption}
\usepackage{hhline}
\usepackage{amssymb}
\usepackage[switch]{lineno}
\usepackage{cite}

\ifCLASSINFOpdf
\else
\fi
\begin{document}
%
\title{Residual Q-Networks for Value Function Factorizing in Multi-Agent Reinforcement Learning }
%
%
%

\author{Rafael~Pina, Varuna~De~Silva, Joosep~Hook
        and~Ahmet~Kondoz,~\IEEEmembership{Senior Member,~IEEE}}
\maketitle

\begin{abstract}
Multi-Agent Reinforcement Learning (MARL) is useful in many problems that require the cooperation and coordination of multiple agents. Learning optimal policies using reinforcement learning in a multi-agent setting can be very difficult as the number of agents increases. Recent solutions such as Value Decomposition Networks (VDN), QMIX, QTRAN and QPLEX adhere to the centralized training and decentralized execution scheme and perform factorization of the joint action-value functions. However, these methods still suffer from  increased environmental complexity, and at times fail to converge in a stable manner. We propose a novel concept of Residual Q-Networks (RQNs) for MARL, which learns to transform the individual Q-value trajectories in a way that preserves the Individual-Global-Max criteria (IGM), but is more robust in factorizing action-value functions. The RQN acts as an auxiliary network that accelerates convergence and will become obsolete as the agents reach the training objectives. The performance of the proposed method is compared against several state-of-the-art techniques such as QPLEX, QMIX, QTRAN and VDN, in a range of multi-agent cooperative tasks. The results illustrate that the proposed method, in general, converges faster, with increased stability and shows robust performance in a wider family of environments. The improvements in results are more prominent in environments with severe punishments for non-cooperative behaviours and especially in the absence of complete state information during training time.
\end{abstract}


%
\IEEEpeerreviewmaketitle

\section{Introduction}
%
%
%
%
\IEEEPARstart{R}{einforcement} learning has been growing as one of the most prominent areas of artificial intelligence in the recent past \cite{chen_new_2019}. It has proved to be successful in single-agent tasks in diverse areas of application such as finance, games, and manufacturing \cite{qu_value_2019, hook1}. There are many applications that require reinforcement learning to be done in Multi-Agent settings, which require some level of cooperation and coordination among the agents \cite{iqbal_actor-attention-critic_2019}.

In centralized Multi-Agent Reinforcement Learning (MARL), observations and actions of all agents are considered as a whole by a centralized oracle. However, centralised learning suffers from Curse of Dimensionality as the number of agents increases and the action space becomes extremely large \cite{nguyen_deep_2020, automata}. On the other end of the spectrum, fully decentralized MARL treats each agent as an independent learner. However, learning in such a setting becomes non-stationary from the perspective of a single agent, and renders it difficult for an agent to understand the reasons for the joint reward. As a result, the agent might get confounded by the rewards of other agents. Thus, the optimal policy may not be learnable in such settings. This problem is referred to as relative overgeneralization \cite{wei_multiagent_2018}. To minimize the above problems, a new paradigm has been investigated for MARL, known as the centralized training and decentralized execution (CTDE). It allows the centralized critic to use the full information available of the environment during training, but restricts agents (actors) to their local observations when it comes to execution. Most of the MARL methods proposed in the recent past focus on the centralized training decentralized execution scheme \cite{lowe_multi-agent_2017, foerster_counterfactual_2017,gupta_2017, hook_multi_critic}. This paradigm helps to minimize the relative overgeneralization problem by improving the credit assignment to the agents, which helps agents to appropriately deduce their individual contribution to the overall team reward \cite{foerster_counterfactual_2017}. Yet, a fully centralized critic still suffers from Curse of Dimensionality \cite{rashid_qmix_2018}. 

\begin{figure}\label{mini_arch}
    \centering
    \includegraphics[width=0.45\columnwidth]{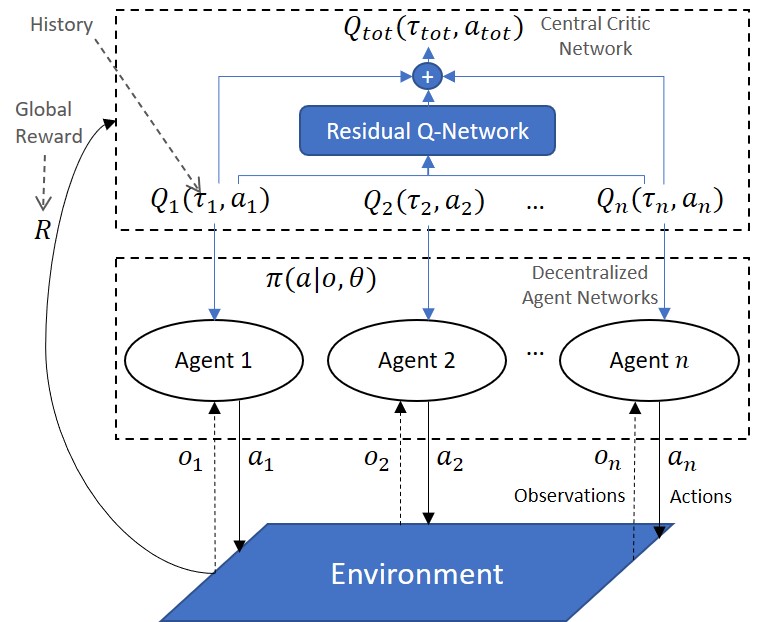}
    \caption{Illustration of the centralized training decentralized execution paradigm within the proposed RQN method.}
    \label{fig:mini_arch}
\end{figure}

One possible approach to overcome the Curse of Dimensionality is to learn a centralized, but a factorized critic, which learns to decompose the joint action-value function to individual action-value functions of the agents. The Value Decomposition Networks (VDN) \cite{sunehag_value-decomposition_2017} and QMIX \cite{rashid_qmix_2018}, are two examples of such techniques. VDN aims to learn an additive decomposition of the joint action-value function. However, authors in \cite{rashid_qmix_2018} argue that the simple linear operations performed by VDN limit the complexity of the value functions that it can represent. QMIX tries to extend the range of value functions that can be represented by mixing the individual Q-values in a monotonic way. However, VDN and QMIX are limited by additivity and monotonicity constraints, respectively, and are unable to factorize action-value functions that do not follow these constraints. \cite{son_qtran_2019} proposed QTRAN, a solution capable of overcoming these problems by using a better factorization process. The main idea of QTRAN is to transform the original action-value function into a new one, and factorize the transformed function. Despite the many solutions brought by these methods, there are still problems to tackle that have been mitigated but not totally overcome yet such as the credit assignment problem \cite{lansdell_learning_2020}.

In this paper we show that there exists a more relaxed condition than in QTRAN \cite{son_qtran_2019} under which value function factorization is possible. Accordingly, we propose a new deep learning based method that aims to learn factorizations of the action-value functions in different tasks by performing complex non-linear operations over the individual Q-values. We propose the use of a new Residual Q-Network (RQN) (Figure \ref{fig:mini_arch}) that will allow to reduce the impact of the credit assignment problem \cite{lansdell_learning_2020,yang_cm3_2020}. In order to achieve that, the RQN will learn an individual factor for each agent that will indicate the relative importance of a certain Q-value trajectory (the sequence of action-values of a given episode). As a result, the proposed method is able to also improve the credit assignment of the agents and achieve a quick and stable level of optimal performance over time by learning estimations for the individual Q-values. The main role of the RQN is to learn estimations for the individual Q-values in order to coax the agents towards optimal performances. The intuition is that if individual agents are performing well by visiting certain states with relatively high Q-values, they should be encouraged to visit those states in subsequent episodes. To do that, the proposed model should take advantage of the individual Q-value trajectories. The empirical results suggest that the proposed approach is affected less by increasing the number of agents, when compared to previous value decomposition-based methods. The performance of the proposed method is benchmarked against previous related approaches (QMIX \cite{rashid_qmix_2018}, QTRAN \cite{son_qtran_2019}, VDN \cite{sunehag_value-decomposition_2017}, QPLEX \cite{wang2020qplex} and WQMIX \cite{rashid_weighted_2020}) in a set of cooperative multi-agent environments of varying complexity.
Contributions of this paper are as follows:
\begin{enumerate}[]
\item We present novel theoretical evidence for factorization conditions in a wider family of different environments. 
\item We propose a new method for value function factorization in Multi-Agent Reinforcement Learning cooperative tasks that uses a novel neural network architecture based on a novel concept of Residual Q-Networks (RQNs).
\item We demonstrate the viability of the proposed method in a diverse set of environments with varying levels of complexity, with distinct tasks and number of agents. 
\end{enumerate}

The rest of the paper is organized as follows: In section \ref{sec:relw} we present recent literature that is associated with the contributions of this paper, and section \ref{sec:bkg} presents the theoretical details fundamental to the developments of this paper. Section \ref{sec:rqn} presents the proposed novel value factorization methodology for MARL tasks. In section \ref{sec:exp}, experimental details are presented, followed by results and discussions in section \ref{sec:res}. We conclude the paper in section \ref{sec:conc} with indications for future work. 
 

\section{Related Work}\label{sec:relw}
\cite{tan_minudletip-aengdeennttrvesincfoorocpemereantitvle_1993} introduced one of the first Multi-Agent Reinforcement Learning approaches, based on independent Q-learning \cite{watkins_1992_technical}. In a more complex but related approach, \cite{tampuu_multiagent_2015} extended the concept with Deep Q-Networks to multi-agent environments, where individual agents controlled by Independent Deep Q-Networks learn both cooperative and competitive behaviours.

Deep reinforcement learning for multi-agent settings quickly became popular, but the increase of the number of agents proved to be a difficult problem to solve. To mitigate this problem, a recurring approach in recent works is the use of a centralized training and decentralized execution model \cite{nguyen_deep_2020}. COMA \cite{foerster_counterfactual_2017} is an example of an actor-critic centralized training decentralized execution method that uses a centralized critic to estimate values for state-action pairs and a decentralized actor that maps states to actions. Other recent works also follow this paradigm such as \cite{lowe_multi-agent_2017}, \cite{yang_cm3_2020} or \cite{rashid_weighted_2020}. Methods such as QMIX \cite{rashid_qmix_2018}, VDN \cite{sunehag_value-decomposition_2017}, QTRAN \cite{son_qtran_2019} and QPLEX \cite{wang2020qplex} also adhere to the centralized training and decentralized execution approach. Moreover, they aim to learn a factorization for the centralized critic, which proved to be a successful strategy. Further, these methods employ a parameter sharing approach to improve learning efficiency \cite{gupta_2017}. However, VDN and QMIX are constrained by additivity and monotonicity, respectively \cite{son_qtran_2019}. In addition, methods like QMIX, QPLEX or QTRAN take advantage of the global state of the environment, which can be unfeasible under certain conditions. On the side, works such as \cite{task_decomp} use similar approaches together with an idea of task decomposition to improve the learning performance.

Communication-based approaches are also popular in multi-agent scenarios \cite{lowe_multi-agent_2017, qu_value_2019,foerster_learning_2016,back_prop}. A key difference of this family of algorithms from fully decentralized execution, is that the agents can communicate among themselves during the execution process. Multiple different works proved that this model can be effective. \cite{das_tarmac_2020} proposes an algorithm where the agents learn to coordinate by performing multiple rounds of communication among themselves before taking an action in the environment. Also, \cite{foerster_learning_2016} demonstrate interesting results by showing that a group of agents is capable of learning complex communication protocols during centralized training phase, to solve cooperative tasks with communication during execution time.

The proposed work in this paper is a new method that follows a centralized training and decentralized execution model for cooperative scenarios. Accordingly, it falls in the category of VDN, QMIX, QTRAN and QPLEX where an effective decomposition of the joint action-value function is learnt. In section \ref{sec:bkg} we elaborate on the theoretical details of these methods.


\section{Background}\label{sec:bkg}
We consider a set of fully cooperative tasks that can be represented as Decentralized Partially Observable Markov Decision Processes (Dec-POMDP) \cite{oliehoeka_2015_a}. A Dec-POMDP is an extension of a Markov Decision Process to a multi-agent setting where the environment is not fully perceptible by each agent. Let the tuple $G\mathrm{=}\left\langle S,A,R,O,Z,P,N\mathrm{,\ }\gamma \right\rangle$ be the representation of a Dec-POMDP, where $S$ is a set of states $s$ that represent the current state of the environment. For each agent $i\in \mathcal{N}\mathrm{\equiv }\mathrm{\{}\mathrm{1,\dots ,}N\}$ consider $A$ to be the set of actions $a_i$ that each agent can choose at a given time step $t$, $a_i\mathrm{\in }A\ \mathrm{:}\ A\mathrm{=}\mathrm{\{}a_{\mathrm{1}}\mathrm{,\dots ,}a_n\}$ and \hl{$R(s,a):S\times A\rightarrow \mathbb{R}$} and \hl{$O(s,i):S \times \mathcal{N} \rightarrow Z$} are the reward and observation functions, respectively. The reward function is shared by all the agents and $Z$ represents the model of observations. The discount factor of the reward function is represented by $\gamma \ :\ \gamma \in \left[0,1\right]$. Since we consider a partially observable system, each agent receives a local observation $o_i$ from the model $O$. Furthermore, each agent $i$ holds an action-observation history represented by ${\tau }_i$ so that \hl{${\tau }_i\mathrm{\in }T\equiv(Z\times A)^*\rightarrow\mathrm{\{}{\tau }_{\mathrm{1}}\mathrm{,\dots ,\ }{\tau }_N\}$.} A transition of state in the environment is performed according to the probability function \hl{$P\mathrm{(}s{\mathrm{'}}\mathrm{|}s,a\mathrm{)}:S \times A \times S \rightarrow [0,1]$,} where $s{\mathrm{'}}$ represents the state that follows $s$ after taking an action $a$. This constructs a certain policy $\pi_i(a_i|\tau_i)$. The objective is to \hl{find an optimal joint policy $\pi$} that maximizes the joint action-value function $Q_{\pi}(s_t,a_t)=\mathbb{E}_{\pi}[R_t|s_t,a_t]$, where $R_t = \sum_{k=0}^\infty \gamma^kR_{t+k}$ is the discounted return.

In this section, we introduce the key concepts to understand the method proposed in this paper.

\subsection{Reinforcement Learning with Deep Q-Networks}
In the work of \cite{mnih_2015_humanlevel}, Deep Q-Networks (DQN) is an approach that uses deep neural networks to approximate value-functions $Q(s,a)$. This is done using a target network, whose parameters are copied every $n$ steps from an evaluation network, where $n$ is a pre-defined number of steps to update the target network. Furthermore, the networks have access to a buffer with an observation-action history used to train them. In Deep Q-Learning the loss of the networks is calculated according to \begin{equation}\label{eq:1}
\mathcal{L}\left(\theta \right)={[y^{DQN}-Q\left(s,a;\theta \right)]}^2 
\end{equation}
where $y^{DQN}$ is the DQN target network, $y^{DQN}=R(s,a)+\gamma {\mathop{\mathrm{max}}_{a'} Q(s',a';{\theta }^-)\ }$,  and $\theta $ and ${\theta }^-$ represent the parameters of the evaluation network and target, respectively.

\subsection{Value Function Factorization in Centralized Training-Decentralized Execution Based MARL}\label{sec:vdn}
Under the centralized training decentralized execution paradigm, recent studies defined a key condition that should be satisfied to enable efficient learning during the centralized training phase. This condition is called Individual-Global-Max (IGM) \cite{son_qtran_2019} and is defined as
\begin{equation}\label{eq:2}
\argmax_aQ_{tot}\left(\tau,a\right)=
\begin{pmatrix}
\argmax_{a_1}Q_1({\tau }_1,a_1) \\
\vdots\\
\argmax_{a_N}Q_N({\tau }_N,a_N)
\end{pmatrix}
\end{equation}
In simple terms, we can describe this property as requiring the optimal joint action to be the same as the combination of locally optimal actions of individual agents.

Value Decomposition Networks \cite{sunehag_value-decomposition_2017} is a method that aims to learn an additive decomposition over individual agents based on the team reward. The key idea is the assumption that a joint action-value function of a group of agents can be represented as the addition of individual action-value functions of each agent. In simple terms, this method decomposes the action-value function according to \begin{equation}\label{eq:3}
    Q_{tot}\left(\tau ,a\right)\mathrm{=}\sum^N_{i\mathrm{=1}}{Q_i\mathrm{(}{\tau }_i,a_i\mathrm{;}{\theta }_i\mathrm{)}}
\end{equation}
where $Q_{tot}$ is the joint action-value function and $Q_i$ represents the individual value functions for each agent $i$ with parameters ${\theta }_i$. However, it is argued that this type of decomposition is too simple to represent a wide range of environments due to the additivity constraint \cite{rashid_qmix_2018,son_qtran_2019}.

QMIX is a method that learns a more complex decomposition of the joint action-value function using hyper networks that take advantage of extra state information during centralized training \cite{rashid_qmix_2018}. The decomposition is done using a mixing network that uses complex non-linear operations that follow an imposed monotonicity constraint, represented as
\begin{equation}\label{eq:4}
    \frac{\partial Q_{tot}}{\partial Q_i}\ge 0,\ \forall i
\end{equation}

The motivation for this monotonicity constraint is to generalize the family of functions that VDN can represent to a larger family of monotonic functions. This condition is assured by keeping the weights of the mixing network positive. This constraint has proved to be a problem when it comes to solving tasks that do not meet the monotonicity conditions \cite{son_qtran_2019,rashid_weighted_2020}. For example, a task where the best action of one agent depends on the actions of a second agent, at the same time step. We demonstrate this effect using a matrix game in subsection \ref{sec:rep_comp}. To update their networks, both VDN and QMIX use the DQN loss as stated in (\ref{eq:1}), but their losses operate on $Q_{tot}$ instead of an individual $Q$ of DQN.

QTRAN \cite{son_qtran_2019} is a different value-decomposition approach that proposes a new method of factorizing joint action-value functions without the constraints imposed by VDN and QMIX: additivity and monotonicity. The key idea of QTRAN is to transform the original action-value function using an additive decomposition, and then to correct it by using a joint state-dependent value function. To address the additivity and monotonicity constraints present in VDN and QMIX, QTRAN proposes a new approach that uses multiple networks that are trained in a centralized way. QTRAN uses an individual action-value network for each agent $i$, $f_q\mathrm{:(}{\tau }_i,a_i\mathrm{)}\mathrm{\mapsto }Q_i$, a joint action-value network, $f_r\mathrm{:(}\tau ,a\mathrm{)}\mathrm{\mapsto }Q_{tot}$, and a state-value network, $f_v\mathrm{:}\tau \mathrm{\mapsto }V_{tot}$. In this sense, QTRAN uses a weighted loss that aims to both learn the optimal $Q_{tot}$ and factorize the transformed action-value function. The use of an unconstrained joint action-value function is key to keep a wide representational complexity. With the method proposed in this paper, we also intend to eliminate both the additivity and the monotonicity constraints by adding more layers of complexity to the decomposition procedure. This process will be explained in the sections ahead.

\subsection{Value Mapping in Residual Networks}\label{sec:resnets}
ResNets was proposed by \cite{he_deep_2015} as a novel approach for image recognition. The key to these networks is the existence of skip connections that allow values to skip some layers in the network. Similarly the work in this paper introduces a novel architecture for MARL with the existence of skip connections. The works of \cite{he_deep_2015} show the strength of such architecture that proves to be easier to optimize than previous networks.

In a Residual Network, the computations performed to map subsequent values are obtained through the equation \cite{he_identity_2016}
\begin{equation}\label{eq:5}
    x_U=x_u+\sum^{U-1}_{i=u}{F(x_i,w_i)}
\end{equation}
where $F$ is a non-linear function with inputs $x$, parameterized by $w$, and $x_U$ are the features of a deeper unit $U$ and $x_u$ are the features of a shallower unit $u$.

\section{Residual Q-Network (RQN) for Value Function Factorization}\label{sec:rqn}
The proposed method involves a centralized, but a factorized critic network supplemented by a novel Residual Q-Network (RQN). In this section we show a more relaxed factorization condition that suits a wider family of joint action-value functions. With the proposed new network architecture, we aim to exploit this condition.

\begin{figure*}\label{rqn_arch}
    \centering
    \includegraphics[height=6cm]{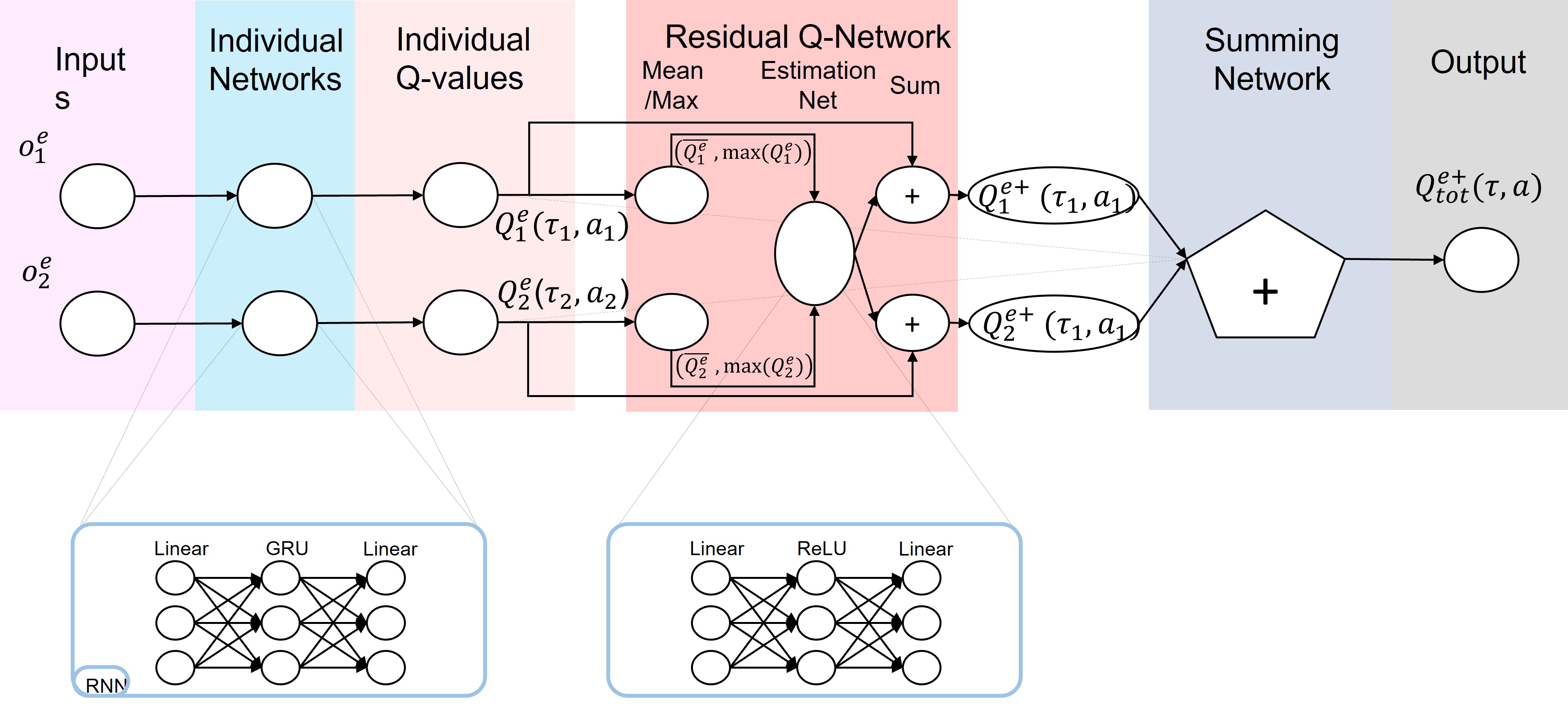}
    \caption{Architecture of the centralized critic where the RQN is described in detail considering an example with two agents, 1 and 2, and for a certain episode $e$.}
    \label{fig:rqn_arch}
\end{figure*}
 
\subsection{Requirements for Factorization with Individual Factor Functions}
We start by defining a Theorem over the Theorem 1 of value function factorization in \cite{son_qtran_2019}, which defines sufficient conditions to satisfy IGM in (\ref{eq:2}).

\begin{theorem}{}
Given a joint action-value function $Q_{tot}(\tau,a)$, we say that it is factorized by $[Q_i(\tau_i,a_i)]$ if
\begin{subnumcases}{\sum^N_{i=1}(Q_i(\tau_i,a_i)+\phi_i(\tau))-Q_{tot}(\tau,a)=}
            0         & $a = \overline{a},$ \label{eq:lemma_eq1} \\
    \geq    0     & $a \neq \overline{a},$ \label{eq:lemma_eq2}
\end{subnumcases}
where $\phi_i$ are the individual correction factors and $\sum^N_{i=1}\phi_i(\tau)={\mathrm{max}}_{a}Q_{tot}(\tau,a)-\sum^N_{i=1}Q_i(\tau_i,\overline{a}_i)$. 
\end{theorem}

Accordingly, we utilize a  transformation of individual action-value functions that preserve IGM, $\sum^N_{i=1}(Q_i(\tau_i,a_i)+\phi_i(\tau))$. This is an affine transformation of the individual factor functions, under which the IGM condition is not violated. This is a much stricter condition than additivity constraint of VDN and monotonicity constraint of QMIX \cite{son_qtran_2019}. Note that this transformation also shares some theoretical similarities with the individual dueling transformation introduced in QPLEX \cite{wang2020qplex}. However, RQN learns a trajectory-based factor and uses this factor to transform the individual $Q_i$ while in the transformation of QPLEX, the authors use advantage functions together with the state-value functions.

\textit{Proof.} Theorem 1 shows that if the above condition holds, then $Q_i$ satisfies IGM. In a similar manner to \cite{son_qtran_2019}, we show that if there is a $Q_i$ that satisfies the conditions above, then the joint action that maximizes the $Q_{tot}$ is the same as the set of optimal local actions $\overline{a}=[\overline{a}_i]^N_{i=1}$, where $a_i$ denotes an optimal local action of the agent $i$, $\overline{a}_i=\argmax_{a_i}Q_i(\tau_i,a_i)$. We have that\\
\begin{align*}\nonumber
    Q_{tot}(\tau, \overline{a})&=\sum^N_{i=1}(Q_i(\tau_i,\overline{a}_i)+\phi_i(\tau))\quad\quad\quad from \ (\ref{eq:lemma_eq1})\\
    &=\sum^N_{i=1}(Q_i(\tau_i,\overline{a}_i))+\sum^N_{i=1}\phi_i(\tau)\\
    &\geq\sum^N_{i=1}(Q_i(\tau_i,a_i))+\sum^N_{i=1}\phi_i(\tau)\\
    &\geq Q_{tot}(\tau,a) \quad\quad\quad\quad\quad\quad\quad\quad\quad from \ (\ref{eq:lemma_eq2})
\end{align*}

This means that the set of optimal local actions $\overline{a}$ maximizes $Q_{tot}$, i.e., $Q_i$ satisfies IGM. End of Proof. \qed

\subsection{Factorizing with Residual Q-Networks}

In this section, we propose a deep reinforcement learning method that exploits Theorem 1. Our motivation is to achieve an effective decomposition by learning a new estimation factor that will affine transform the individual Q-value trajectories $Q_i(\tau_i,a_i)$. A  different factor $\phi_i(\tau)$ is learnt for each agent, so that we can prioritize different state-action trajectories taken by agents within an episode. 

The proposed architecture for the centralized critic is illustrated in Figure \ref{fig:rqn_arch}. The individual agent networks are recurrent neural networks using a GRU with width 64. These networks receive the observations $o_i$ from the agents at a given time step and will output individual action-values based on the observation-action histories. These values are passed on to the RQN, where a set of non-linear operations are performed. Over a batch of episodes, an estimation network (within the RQN) will take as input features the mean of the Q-values of each agent over the steps of each episode, and their maximum Q-value. This estimation network has an input shape of $2\times N$, where $N$ is the number of agents. The inputs are given to a Linear layer with 64 units, followed by a ReLU non-linearity and finally another Linear layer, where the output has size equal to $N$. In other words, the output is a set with $N$ estimation factors, one for each agent. The estimation factors are added to the initial action-values from the individual agent networks, to compute the adjusted individual Q-values. The adjusted Q-values are given as inputs to the summing network, to calculate the joint action-value function.

 We denote estimation factor, by which the individual Q-values are adjusted, as ${\phi }_i$ for each agent $i\mathrm{\ :\ }i\mathrm{\in }\mathrm{\{}\mathrm{1,\dots ,}N\}$. This factor is the output of the RQN. Learning this factor successfully is the first aspect of the proposed method. The network is trained by receiving two features for each agent, one being the mean of the Q-values across all the steps $t\ :t\in \left\{1,\dots ,T\right\}$ of each episode $e$,
\begin{equation}\label{eq:7}
  {\overline{x}}^e_i=\frac{\sum^T_{t=0}{Q^t_i(s^t_i,a^t_i)}}{T}  
\end{equation}
and the second is the maximum Q-value of each agent across all the steps of each episode $e$,
\begin{equation}\label{eq:8}
  m^e_i={\mathop{\mathrm{max}}_{t} Q^t_i(s^t_i,a^t_i)\ }  
\end{equation}

The intuition to include the maximum as a feature to the estimation network (see Figure \ref{fig:rqn_arch}) is that if the trajectory of an agent contains a relatively high value state-action pair, it should be encouraged to explore it in subsequent episodes. Further, mean value of a trajectory signals the relative goodness of an entire trajectory and maximum value will signify the goodness of some points within the trajectory.
The estimation factor for each agent and an episode $e$ can be calculated according to 
\begin{equation}\label{eq:9}
  {\phi }_i=f\left([{\overline{x}}^e_i],[m^e_i];{\theta }\right)_i,\ \ ~i\in \left\{1,\dots ,N\right\}  
\end{equation}
where $f$ represents the RQN with parameters $\theta $.
This estimation factor, which is a non-linear function of the action-value trajectories, allows to decompose action-value functions of more complex environments than VDN due to the elimination of the simple sum constraint \cite{rashid_qmix_2018}. Furthermore, the factorization procedure of the proposed method can benefit from the trajectories of the individual Q-values. In contrast to QMIX, the RQN based method removes the monotonicity constraint by not enforcing the weights of the network to remain positive. In comparison to QTRAN, where the state-value network computes a single value for $Q_{tot}$, the RQN calculates an estimation factor for each individual Q-values separately.

After the estimation factor for each agent is calculated by the network, the factor is added to the respective Q-value trajectories  resulting in a new estimated Q-value for each agent $i$, according to $Q^+_i=Q_i+{\mathit{\phi}}_i$, where ${\mathit{\phi}}_i$ is obtained from (\ref{eq:9}). 

At this point, we have learned an estimated version of the Q-values for each agent, the estimated Q-values, denoted by $Q^+_i$. Now, the critic network can be used to compute a new joint action-value function using the estimated Q-values,
\begin{equation}\label{eq:10}
  Q^+_{tot}=\sum^N_{i=1}{Q^+_i~({\tau }_i,a_i;{\theta }_i)}  
\end{equation}

The proposed method is capable of learning a decomposition more capable of factorizing different factorizable tasks than previous methods due to the use of a RQN. As described in (\ref{eq:5}) consider now that $H\left(Q_i\right)=F\left(x\right)+Q_i$ gives the output of the RQN. Recalling (\ref{eq:7}) and (\ref{eq:8}), consider that $x={(\overline{x}}^e_i,m^e_i)$ is a set containing the inputs for an agent $i$ and an episode $e$ to the estimation network of our model (Figure \ref{fig:rqn_arch}). Consider the example where the optimal output of the RQN given this input $Q_i$ would be $H\left(Q_i\right)=Q_i$. This would be difficult to approximate using the normal mapping of a common neural network \cite{he_deep_2015}, since by inputting $Q_i$ then it would need to be optimally mapped to $Q_i$ after going through the network. On the other hand, if we can compute a value $F\left(x\right)$ given the input $x$ as a combination of features regarding the original input to the RQN, and $F$ being the estimation network, then we end up with a factor ${\phi }_i$ (as represented in (\ref{eq:9})) given by $F\left(x\right)={\phi }_i$ that will converge to a stable value when the training objective is reached, since at that point, the role of the RQN as an auxiliary network is no longer needed. We investigate this effect further in subsection \ref{sec:add}. Thus, $H\left(Q_i\right)=F\left(x\right)+Q_i$ will be easier to learn with the RQN than with a common neural network since $F\left(x\right)$ stabilizes and $Q_i$ skips connections as described in \ref{sec:resnets}.

\subsection{Comparison of Representational Capacity of RQN with QTRAN, VDN and QMIX}\label{sec:rep_comp}
For this demonstration, let $Q_{jt}$ represent a joint action-value function. The key idea of QTRAN is to transform the original joint action-value function $Q_{jt}$ into a new one $Q_{jt}^\prime$ that shares the optimal joint action with $Q_{jt}$. Theorem 1 in QTRAN \cite{son_qtran_2019} states a sufficient condition for $[Q_i]$ that satisfies IGM, $i\mathrm{\in }\mathrm{\{}\mathrm{1,\dots ,}N\}$. Accordingly, a factorizable joint action value function $Q_{jt}(\tau,a)$ is factorized by $[Q_i(\tau_i,a_i)]$, if

\begin{subnumcases}{\sum^N_{i=1}Q_i(\tau_i,a_i)-Q_{jt}(\tau,a)+V_{jt}(\tau)=}
            0         & $a = \overline{a},$ \label{eq:comp_eq1} \\
    \geq    0     & $a \neq \overline{a},$ \label{eq:comp_eq2}
\end{subnumcases}
\\
where $V_{jt}={\mathrm{max}}_{a}Q_{jt}(\tau,a)-\sum^N_{i=1}Q_i(\tau_i,\overline{a}_i)$ and $\overline{a}$ is a set of optimal local actions $\overline{a}_i$.\\
\newline
The above condition becomes necessary under an affine transformation of action-value functions, $\psi=A\cdot Q+B$, where $A=[a_{ii}]\in{R}_+^{N\times N}$ is a symmetric diagonal matrix and $B=[b_{i}]\in{R}^{N}$. An important observation is that values of $B$ does not alter the IGM condition. Thus, QTRAN relies on a linear transformation of individual factor functions, $Q_{jt}^\prime(\tau,a)=\sum_{i=1}^{N}{Q_i(\tau_i,a_i)}$, a transformation that contains IGM. Thus, the transformation used in QTRAN is an identity matrix.

The representational power of QTRAN over VDN relies on the $V_{jt}(\tau)$, which tries to learn the difference (or the residual) between the joint action-value function $Q_{jt}(\tau,a)$ and the transformation at individually optimal actions $Q_{jt}^\prime(\tau,\overline{a})$. Thus $V_{jt}(\tau)$, is a residual function, that acts on top of the transformation.

In comparison, the proposed RQN, relies on an affine transformation, on individual factor functions. Thus, giving more flexibility for learning the individual maximums, while retaining the IGM property beyond additivity and monotonicity constraints.

In comparison to QTRAN, the transformation adopted by RQN is as follows, where $\phi_i$ are individual correction factors:
\begin{equation}
    Q_{jt}^\prime=\sum^N_{i=1}(Q_i(\tau_i, a_i) + \phi_i(\tau))
\end{equation}
In comparison to QTRAN, the representational difference with VDN, in RQN comes from $\sum_{i=1}^{N}\phi_i$.

The success of the proposed RQN structure is due to the difference in calculating the residual factors. In QTRAN, the authors rely on the joint action value function $Q_{tot}(\tau,a)$, to learn the residual between transformation, whereas in RQN we rely on a transformation that adheres to the sufficient conditions of IGM as much as QTRAN, but the learning problem is formulated to finding the individual maximums.

To further illustrate the differences in the representational complexities of the different approaches, we show the reconstructions of the $Q_{tot}$ in a one-step matrix game, as used in \cite{son_qtran_2019} and \cite{rashid_weighted_2020}. In this simple game, two agents can choose one of three different actions in order to optimize a joint action, given by $Q_{tot}$. The matrix of the game is a non-monotonic matrix represented by Table \ref{sub_tab:mat}. The agents are trained for 20000 episodes (1-step episodes) in full exploration ($\epsilon = 1$) and with a replay buffer of size 500. Tables \ref{sub_tab:vdn}-\ref{sub_tab:rqn} show the reconstructed $Q_{tot}$ for VDN, QMIX, QTRAN and RQN, respectively. As the tables show, only QTRAN and RQN are capable of sucessfuly reconstructing this non-monotonic matrix, whereas VDN and QMIX fail. This supports the fact that RQN can represent a wider family of action-value functions, eliminating constraints such as monotonicity or additivity. 

\begin{table}[h]
\caption{Payoff matrix of the one-step game and reconstructed $Q_{tot}$ for VDN, QMIX, QTRAN, and RQN.\label{tab:table_mtx}}
\centering
\begin{subtable}[h]{\columnwidth}
\centering
\resizebox{!}{18pt}{
\begin{tabular}{|c||c|c|c|}
\hhline{-||---}
$a_0$\textbackslash $a_1$ & A & B & C\\
\hhline{=::===}
A & 8 & -12 & -12\\
\hhline{-||---}
B & -12 & 0 & 0\\
\hhline{-||---}
C & -12 & 0 & 0\\
\hhline{-||---}
\end{tabular}
}
\caption{Payoff matrix}\label{sub_tab:mat}
\end{subtable}

\bigskip

\begin{subtable}[h]{0.45\columnwidth}
\centering
\resizebox{!}{18pt}{
\begin{tabular}{|c||c|c|c|}
\hhline{-||---}
$a_0$\textbackslash $a_1$ & A & B & C\\
\hhline{=::===}
A & -7.50 & -5.72 & -5.18\\
\hhline{-||---}
B & -5.70 & -3.91 & -3.37\\
\hhline{-||---}
C & -4.97 & -3.19 & -2.65\\
\hhline{-||---}
\end{tabular}
}
\caption{VDN}\label{sub_tab:vdn}
\end{subtable}
\begin{subtable}[h]{0.45\columnwidth}
\centering
\resizebox{!}{18pt}{
\begin{tabular}{|c||c|c|c|}
\hhline{-||---}
$a_0$\textbackslash $a_1$ & A & B & C\\
\hhline{=::===}
A & -7.55 & -7.55 & -7.55\\
\hhline{-||---}
B & -7.55 & -0.01 & -0.01\\
\hhline{-||---}
C & -7.55 & -0.01 & -0.01\\
\hhline{-||---}
\end{tabular}
}
\caption{QMIX}\label{sub_tab:qmix}
\end{subtable}

\bigskip

\begin{subtable}[h]{0.45\columnwidth}
\centering
\resizebox{!}{18pt}{
\begin{tabular}{|c||c|c|c|}
\hhline{-||---}
$a_0$\textbackslash $a_1$ & A & B & C\\
\hhline{=::===}
A & 7.97 & -12.00 & -11.99\\
\hhline{-||---}
B & -12.00 & 0.00 & 0.00\\
\hhline{-||---}
C & -11.99 & 0.00 & 0.00\\
\hhline{-||---}
\end{tabular}
}
\caption{QTRAN}\label{sub_tab:qtran}
\end{subtable}
\begin{subtable}[h]{0.45\columnwidth}
\centering
\resizebox{!}{18pt}{
\begin{tabular}{|c||c|c|c|}
\hhline{-||---}
$a_0$\textbackslash $a_1$ & A & B & C\\
\hhline{=::===}
A & 8.00 & -12.00 & -12.00\\
\hhline{-||---}
B & -11.99 & 0.00 & 0.00\\
\hhline{-||---}
C & -11.99 & 0.00 & 0.00\\
\hhline{-||---}
\end{tabular}
}
\caption{RQN}\label{sub_tab:rqn}
\end{subtable}
\end{table}

\subsection{Training the Residual Q-Network}
Similar to the description of DQN as in (\ref{eq:1}), the networks that we use during learning are trained to minimize the TD loss that is now generalized for $Q^+_{tot}$ as
\begin{equation}\label{eq:11}
  {\mathcal{L}}_b\left({\theta }_b\right)=\mathbb{E}\left[{(y^{tot}_b-Q^+_{tot}\left(\tau ,a;\theta \right))}^2\right],\ \ b\in \{1,\dots ,B\}  
\end{equation}
where $B$ is the number of batches sampled from the replay buffer and $y^{tot}_b=R+\gamma {\mathop{\mathrm{max}}_{a'} Q_{tot}({\tau }',a';{\theta }^-)\mathrm{\ }\ }$is the target for an iteration $b$, where ${\theta }^-$ represents the parameters of the target network. This loss is propagated firstly through the entire evaluation network and then through the RQN. 
\begin{figure}\label{maps}
    \centering
    \includegraphics[width=\columnwidth]{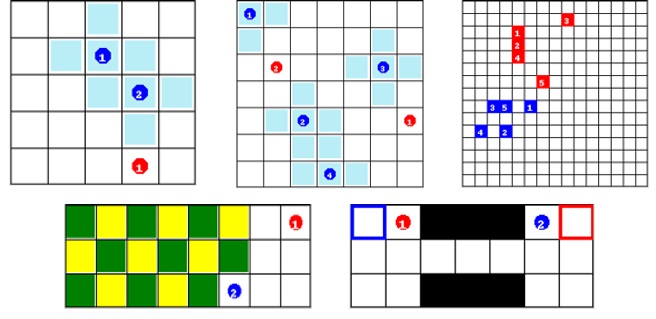}
    \caption{Representation of the matrix environments used in the experiments \cite{magym}. On the top, from the left to the right: PredatorPrey\_2, PredatorPrey\_4, Combat. On the bottom, from the left to the right: Checkers, Switch.}
    \label{fig:maps}
\end{figure}
\begin{figure}\label{maps_smac}
    \centering
    \includegraphics[width=\columnwidth]{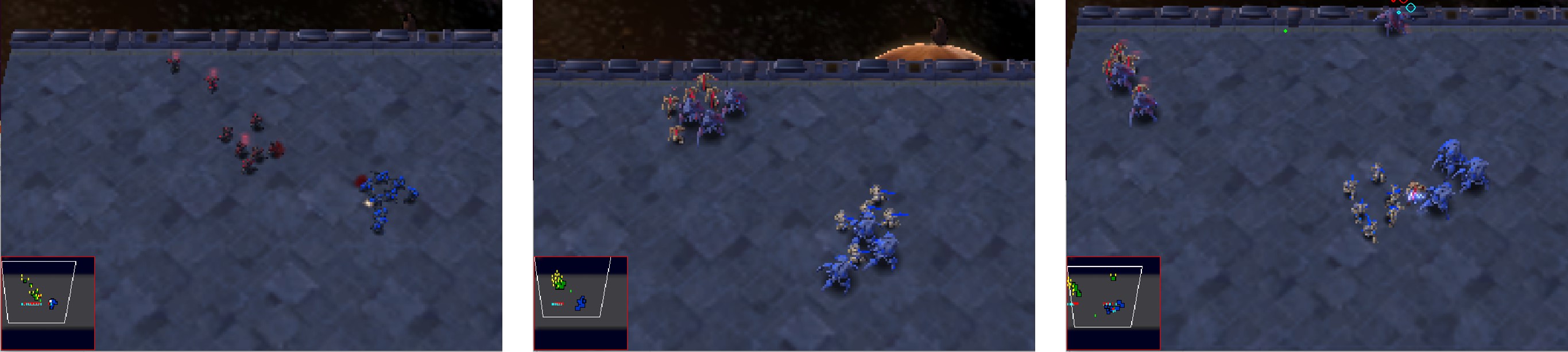}
    \caption{Representation of the Starcraft II environments used in the experiments. From the left to the right: \textit{8m}, \textit{3s5z}, \textit{3s5z\_vs\_3s6z}.}
    \label{fig:maps_smac}
\end{figure}
As such, the targets and the evaluation values are calculated in a slightly different way. The target network is based on $Q_{tot}$ instead of $Q^+_{tot}$. In other words, the target network of the proposed architecture is based on the output of a VDN network, $Q_{tot}$, as described previously in subsection \ref{sec:vdn}. The motivation for this is that by keeping a separate set of targets, we can approximate them with successively minimized losses due to our estimation factor. The RQN will eventually reach a state where it is no more needed since the Q-values will be optimal and stable. At that point, it will render itself obsolete. (this is a key difference of the proposed RQN from ResNets \cite{he_deep_2015}). Thus, the RQN acts as an auxiliary acceleration network that will encourage the agents to converge quickly, by aiding the decomposition of the joint action-value function.

\section{Experimental Details\protect\footnote{Code at \href{https://github.com/rafaelmp2/residual-q-net}{https://github.com/rafaelmp2/residual-q-net}}}\label{sec:exp}
\subsection{Details on the Environments Used}
The performance of the proposed method is evaluated in four different matrix environments with distinct objectives and Starcraft II (SC2) \cite{samvelyan19smac} (Figure \ref{fig:maps_smac}). The maps of the first set of environments are 2D matrices (Figure \ref{fig:maps}) where the action space is discrete, and the observation space is continuous. This set of environments was chosen since similar versions of them were used in previous works such as QTRAN \cite{son_qtran_2019} and VDN \cite{sunehag_value-decomposition_2017}. Note that in the described environments, the individual rewards attributed are then summed up to form a shared team reward. 

\textbf{Switch}: two agents start in opposite cells of the map, separated by a narrow corridor and must switch positions. This environment helps to evaluate how the agents react to a delayed reward scheme. \textbf{Checkers}: on a map with apples and lemons, the agents must eat all the existing apples on the map and avoid lemons. For that, the agents need to cooperate to maximize the team reward. \textbf{PredatorPrey}: a group of predators must catch a group of moving prey. There is a step penalty of -0.01 and every time one of the prey is caught, each agent in the team receives a reward of +5. At least two agents are needed to catch one prey. Two different versions of this environment are used: one where there are 2 predators and 1 prey and in the other, there are 4 predators and 2 prey. These are referred to as PredatorPrey\_2 and PredatorPrey\_4, respectively. Since this environment only has a success reward and no significant intermediate penalties, the task remains monotonic \cite{son_qtran_2019}. \textbf{Combat}: a team of five agents must learn to cooperate to eliminate all the agents of the enemy team. \textbf{SC2}: to complement our experiments we used environments from the challenging Starcraft Multi-Agent Challenge (SMAC) set of environments \cite{samvelyan19smac}. The maps used are the \textit{8m}, \textit{3s5z}, \textit{3s5z\_vs\_3s6z}, all with 8 agents. The level of difficulty was set to very difficult (level 7).
\begin{figure*}[!hbt]\label{plots_1}
    \centering
    \includegraphics[width=\textwidth]{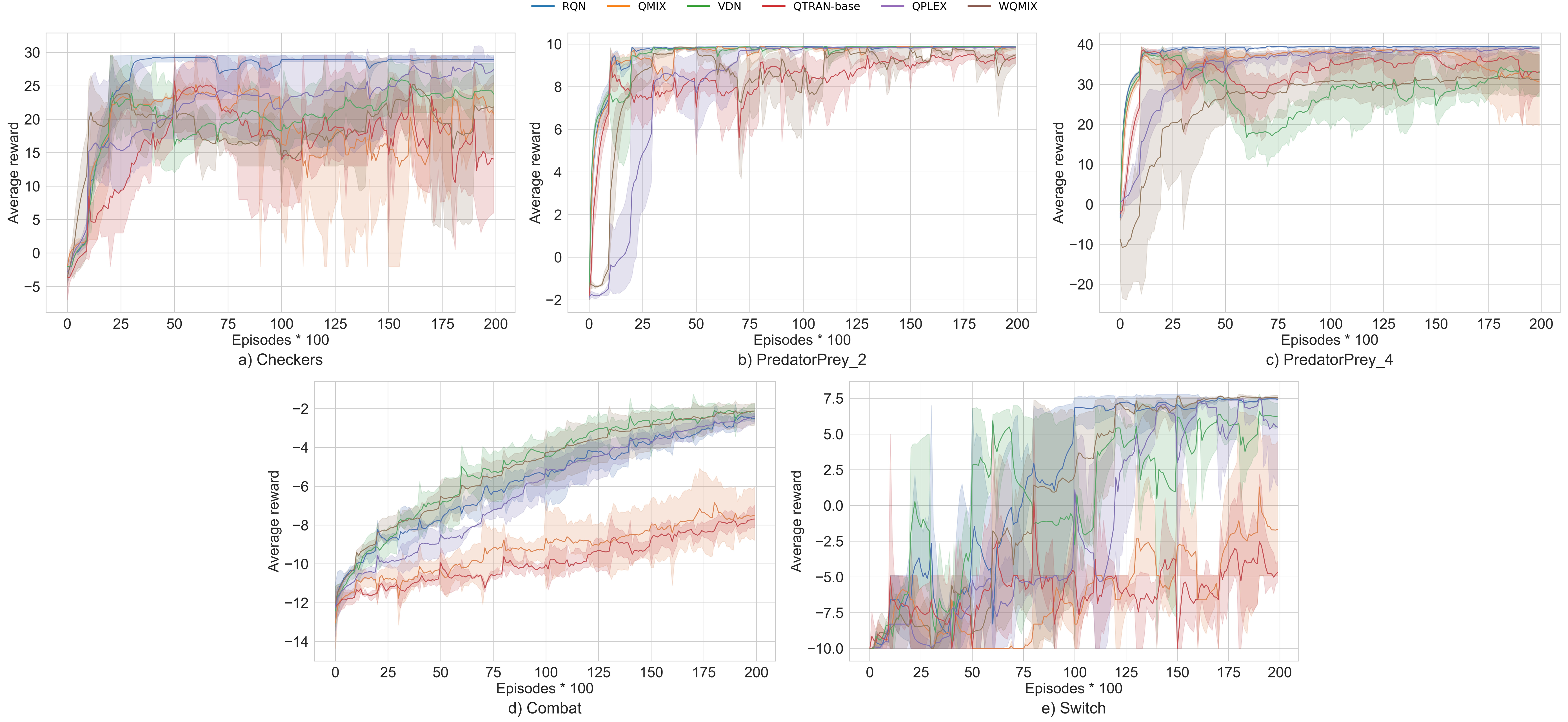}
    \caption{Smoothed average rewards for the experimented methods in the matrix environments. The bold area represents the 95\% confidence intervals.}
    \label{fig:plots_1}
\end{figure*}

\begin{figure*}[!hbt]\label{plots_smac}
    \centering
    \includegraphics[width=\textwidth]{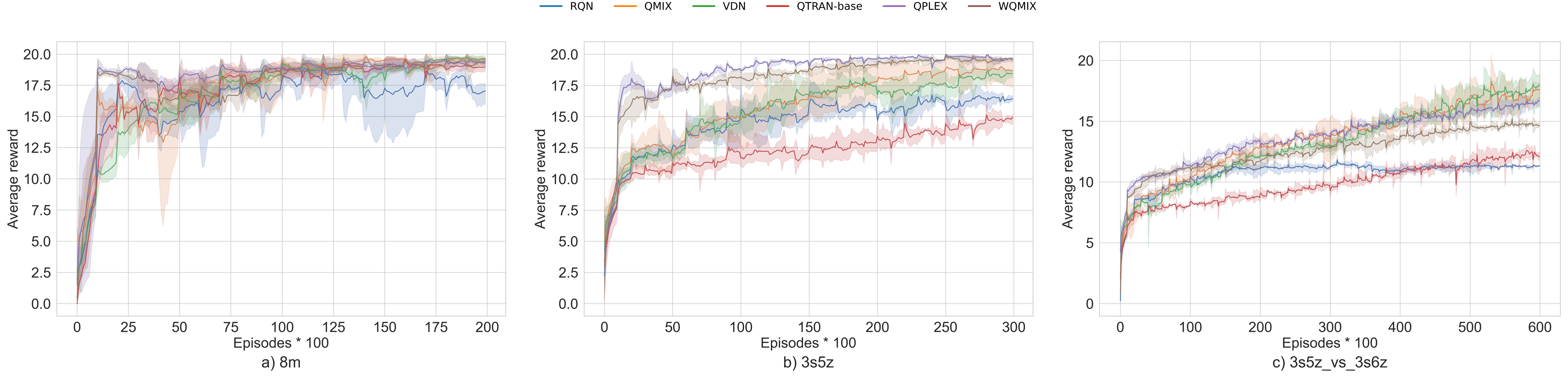}
    \caption{Smoothed average rewards in the attempted Starcraft II environments. The bold area represents the 95\% confidence intervals.}
    \label{fig:plots_smac}
\end{figure*}

\subsection{Algorithms Used as Benchmarks}
The performance of the proposed method is benchmarked against other related approaches using the average episode reward as metric. The agents are trained for a different number of episodes depending on the complexity of the environment, each in 3 independent runs starting at different random seeds. During training time, at every 100 episodes, the model performance is evaluated by running 20 episodes and by averaging the joint rewards for each episode. The proposed approach is compared against VDN \cite{sunehag_value-decomposition_2017}, QMIX \cite{rashid_qmix_2018}, QTRAN-base \cite{son_qtran_2019}, QPLEX \cite{wang2020qplex} and Weighted QMIX (WQMIX) \cite{rashid_weighted_2020}. The width of the hidden layers is 32 for QMIX and 64 for RQN and the other methods, as described in the respective papers. The rest of the hyperparameters used in the main experiments are summarized in \ref{table1}. 

\section{Results and Discussions}\label{sec:res}
In this section we present now the results of the experiments described in the previous section. We start by discussing the results in the matrix environments and analyse the stability of the learning process. Then we discuss the results in the Starcraft II environments and finally we present additional results to analyse the algorithm performance.

\subsection{Performance in Matrix Environments}
Figure \ref{fig:plots_1} illustrates the evolution of the performance of the proposed Residual Q-Network (abbreviated as RQN) and the benchmarked methods. The curves represent the smoothed average rewards over the episodes (smoothed with a 10-step cumulative moving average). For the Switch environment (Figure \ref{fig:plots_1}e), RQN and WQMIX methods manage to solve the task in a small amount of time. Considering that this is a task with delayed rewards, this enforces the intuition that the proposed RQN method encourages the agents to prefer promising states once they are explored. In other words, since the agents do not receive intermediate information about whether their actions are good or not, once they get to the goal, the proposed method encourages them to visit that state again successively, which explains the stability once an optimal state is achieved, making RQN method outperform VDN over time. In the additional experiments subsection it will be clearly illustrated the concept of stability. VDN and QPLEX also solves this task, but are     unable to do it in the same amount of time as the previous, while QMIX and QTRAN-base fail to solve it.

The results in \ref{fig:plots_1}a show that the RQN method can solve a complex non-monotonic task like Checkers successfully. Furthermore, we can see that it is capable of maintaining a stable performance over time once it learns to solve the task. In comparison, although QMIX, VDN, QPLEX QTARN-base and WQMIX manage to solve the task, they fail to achieve as high rewards and stable performance as RQN.

Analysing the performances obtained for two versions of PredatorPrey, all the proposed methods can achieve optimal rewards. In PredatorPrey\_2 (Figure \ref{fig:plots_1}b), all the evaluated methods are able to maintain a stable optimal reward over time. Moreover, since this is a monotonic task, QMIX can solve it successfully as it was expected. However, when the number of agents is increased and hence the complexity of the environment, in PredatorPrey\_4 (Figure \ref{fig:plots_1}c) we observe a different scenario. Despite all the methods being capable of achieving an high value, VDN shows an unexpected behaviour in this task with a notable decrease of performance after around 5000 episodes of training. We believe this happens because the agents get confused due to the exploration activities of the other agents. With the proposed method we aim to solve this problem by improving the credit assignment of the agents and consequently reducing the relative overgeneralization. The results suggest that this problem does not incur in RQN. Surprisingly, QTRAN-base and WQMIX stay slightly below the other methods, while QPLEX also solves this task successfully achieving high rewards. However, RQN can still maintain a higher and more consistent performance. Combat is the exception to the described pattern in the previous environments, where RQN stays below the performance of VDN or WQMIX in the final training steps. On the other hand, QTRAN-base fails to solve this environment and as expected, since this task is not monotonic, QMIX also fails to solve it. 

\subsection{Stability of Performance in the Learning Process}
In Figure \ref{fig:plots_1} we can identify a pattern in all the environments (with the slight exception of Combat) where RQN shows a stable result in the reward over time, opposing to the other methods. Although most of them are able to solve the tasks, they fail to achieve as high values as RQN and show a lack of stability over time, due to high variance of the rewards obtained and to failing to consistently achieve an optimal reward. While RQN is able to maintain the performance achieving high rewards over time, we can see that the other methods keep decaying. In Switch, RQN is also the only one able to maintain a stable performance over time. We call this effect the stability of the learning process. We believe the exceptions with Combat are due to the complexity of the environment allied to its very delayed rewards. Generally, in Figure \ref{fig:plots_1} it is possible to see that RQN displays a very reduced number of oscillations when the convergence (learning goal) was achieved. This is a key difference from the other methods that fail to maintain a stable performance, even when the learning objective is achieved. Observing now PredatorPrey\_4 (Figure \ref{fig:plots_1}c), we can see that QMIX achieves high performance but shows it to be very unstable over time in this four-agent task. This observation suggests that, although the task being monotonic, QMIX fails to scale with the increase in the number of agents. In comparison to QPLEX, although it is capable of maintaining optimal and stable values over time, the proposed RQN method can achieve the same optimal values and, at the same time, shows improved level of stability.

\begin{table*}[ht]
\caption{Hyperparameters used in the experiments in section \ref{sec:res}.}
\label{table1}
\vskip 0.15in
\begin{center}
\begin{small}
\begin{sc}
\begin{tabular}{lcccr}
\toprule
HYPERPARAMETER & VALUE & DESCRIPTION \\
\midrule
batch size    & 32 & number of samples at each training step\\
buffer size   &5000& maximum number of samples to store in the buffer\\
initial $\epsilon$    &1& starting value for $\epsilon$\\
minimum $\epsilon$    &0.05& final value for $\epsilon$\\
$\epsilon$ anneal steps     &50000& number of steps that $\epsilon$ takes to decay\\
\bottomrule
\end{tabular}
\end{sc}
\end{small}
\end{center}
\vskip -0.1in
\end{table*}

\subsection{Performance in Starcraft II Environments}
Figure \ref{fig:plots_smac} illustrates the average rewards in the SC2 environments (smoothed with a 10-step cumulative moving average). The corresponding win rates can be found in the Appendix \ref{appendix}. Although RQN may not outperform the other methods in this set of environments, it shows to be competitive. In the attempted maps, like the majority of the compared methods, RQN can learn the 3s5z and 8m tasks and also show competitive performance in the 3s5z\_vs\_3s6z, although none of them is able to solve the last. Comparing to QTRAN-base RQN can still stay above in the last two tasks. These results, linked to the previous in figures \ref{fig:plots_1}, support the fact that RQN can perform in a wide family of environments due to its relaxed factorization.

Overall, we can see that QTRAN could not learn a good factorization for Combat and Switch environments. QMIX and VDN also fail when monotonicity and additivity are not met, respectively \cite{son_qtran_2019}. When compared to QPLEX and WQMIX, the results suggest that RQN is capable of learning tasks with high performance in a more variate number of environments. More specifically, RQN proves to be competitive in the SC2 environments attempted, linked to a superior performance in the tasks illustrated in Figure \ref{fig:plots_1}. This suggests that RQN can learn a more variate number of different environments, and specially in the absence of extra state information during training. Methods that use a relaxed factorization like RQN and QTRAN might not perform as well in environments such as SC2 \cite{wang2020qplex}. However, a good relaxation proves to be very effective in environments where non-cooperative behaviours are punished more severely \cite{son_qtran_2019}, such as the ones in Figure \ref{fig:plots_1}. The results in Figures \ref{fig:plots_1} and \ref{fig:plots_smac} confirm that, with a more relaxed factorization, RQN can factorize a wider family of environments, even when compared to QTRAN that also uses a relaxed and unconstrained factorization.
\begin{figure}\label{pp2}
    \centering
    \includegraphics[width=\columnwidth]{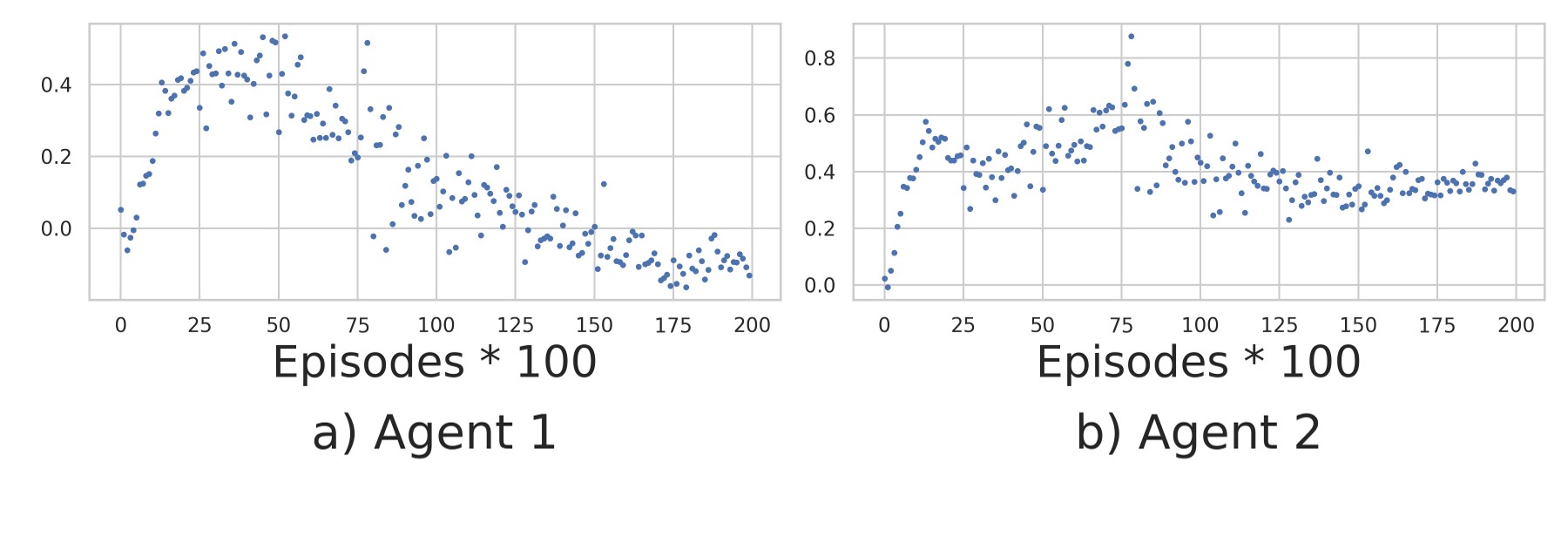}
    \caption{Estimation factors learned by the RQN over time for the two agents in PredatorPrey\_2.}
    \label{fig:pp2}
\end{figure}
\subsection{RQN Self-Depreciation and Task Monotonicity}\label{sec:add}
In this subsection, additional experiments on the algorithm performance are presented to complement the main results in section \ref{sec:res}.

\subsubsection{Estimation Factors Learned by the RQN}

\begin{table*}[ht]
\caption{Hyperparameters used in subsubsection \ref{sec:monot} for the modified PredatorPrey\_4.}
\label{table2}
\vskip 0.15in
\begin{center}
\begin{small}
\begin{sc}
\begin{tabular}{lcccr}
\toprule
HYPERPARAMETER & VALUE & DESCRIPTION \\
\midrule
batch size    & 32 & number of samples at each training step\\
buffer size   &100000& maximum number of samples to store in the buffer\\
initial $\epsilon$    &1& starting value for $\epsilon$\\
minimum $\epsilon$    &0.1& final value for $\epsilon$\\
$\epsilon$ anneal steps     &2000000& number of steps that $\epsilon$ takes to decay\\
\bottomrule
\end{tabular}
\end{sc}
\end{small}
\end{center}
\vskip -0.1in
\end{table*}

Figure \ref{fig:pp2}, illustrates the estimation factors learned by the RQN over time in the PredatorPrey\_2 environment. It is possible to see that it learns an estimation factor over time, and when the performance objectives are reached, the estimation factor for each agent converges to a stable value. Once it reaches the stable value, the purpose of RQN as an auxiliary network that assists in value factorization becomes obsolete. We call this effect the self-depreciation of the RQN when the training objective is reached.

\begin{figure}\label{pp4}
    \centering
    \includegraphics[width=\columnwidth]{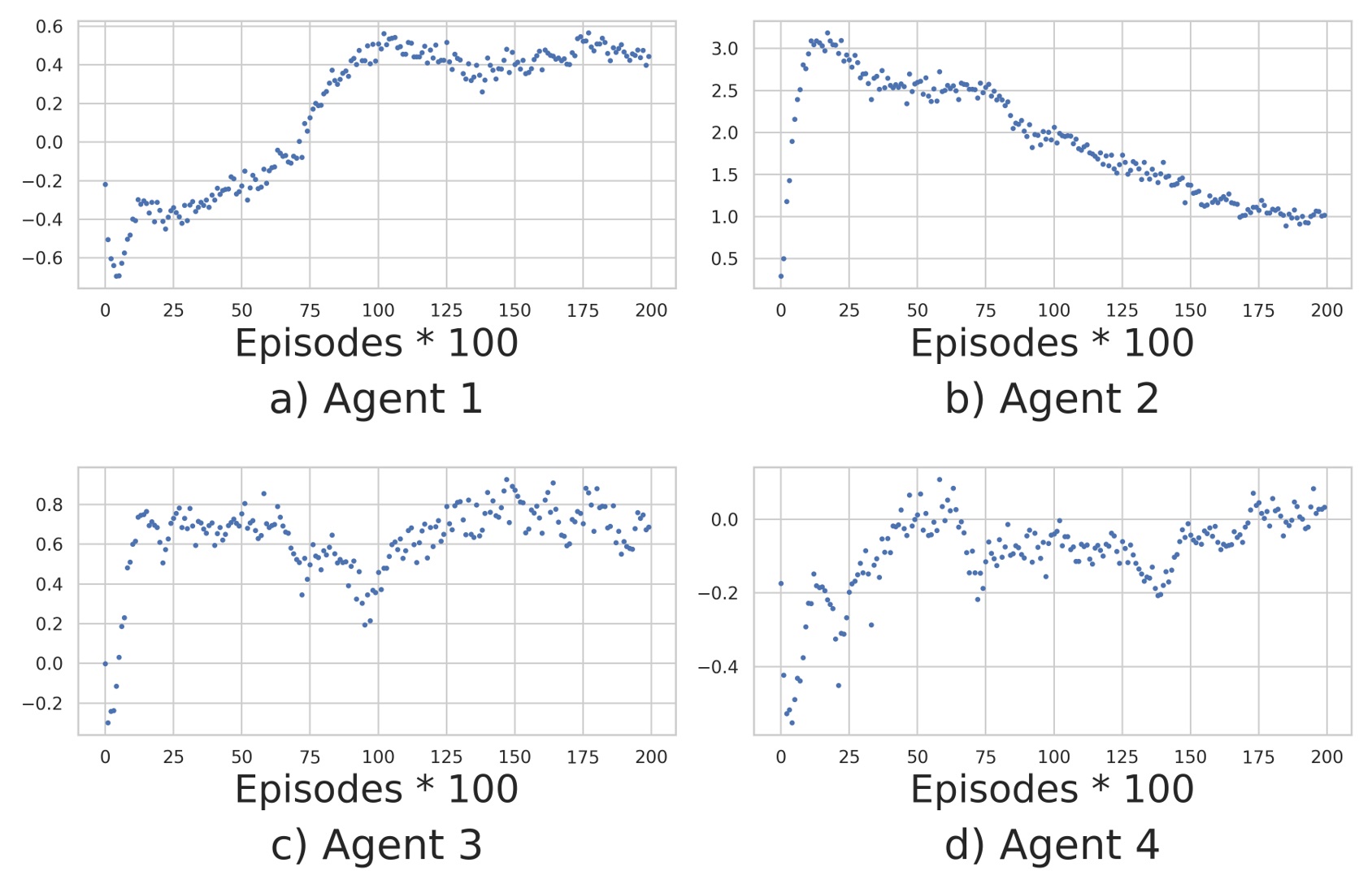}
    \caption{Estimation factors learned by the RQN over time for the four agents in PredatorPrey\_4.}
    \label{fig:pp4}
\end{figure}

Figure \ref{fig:pp4} shows the same metric for the PredatorPrey\_4 environment 
(with 4 agents). This environment is more complex due to the increased number of agents. Thus, although we can see the estimation factor clearly converging for agents 1 and 2, with agent 3 and 4 the convergence has not stabilized. This suggests that, in this case, the agents are still exploring.\\
These representations show how the RQN self-deprecates, because when the training objective is reached, the values estimated by this network should remain stable. The reason is that at the moment of convergence and the objective was achieved, the individual trajectories no longer need to be adjusted.
\begin{figure}\label{pp4_pen}
    \centering
    \includegraphics[width=0.75\columnwidth]{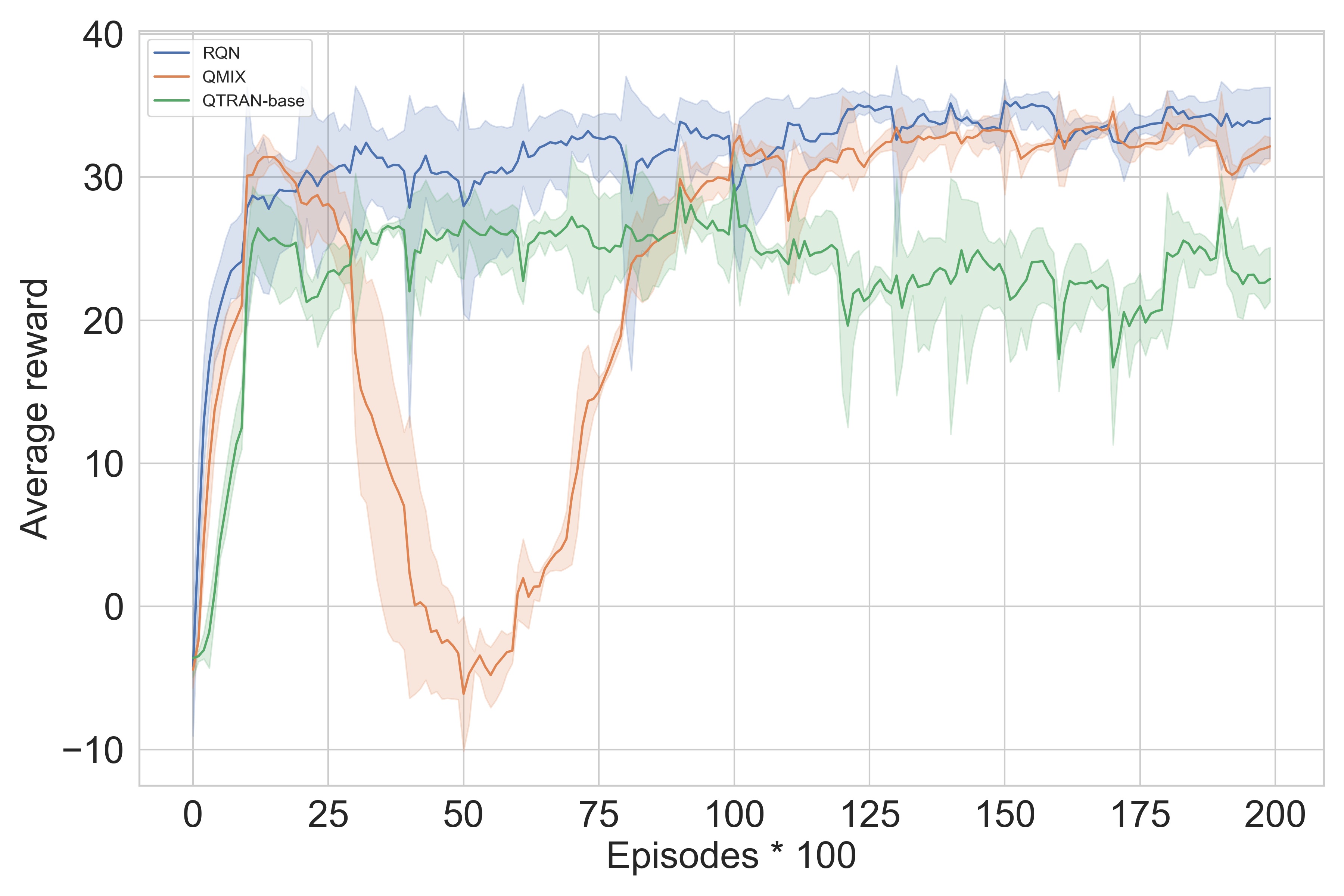}
    \caption{Smoothed average rewards for PredatorPrey\_4. The bold area represents the 95\% confidence intervals.}
    \label{fig:pp4_pen}
\end{figure}
\subsubsection{Impact of a Lower Level of Monotonicity}\label{sec:monot}
To evaluate the performance of QTRAN, QMIX and RQN against a change in the level of the monotonicity in the tasks experimented in this paper, we show now the results for the same PredatorPrey task with 4 agents as in \ref{sec:res} but now with a capture penalty of -0.1. This means that, if one of the agents tries to capture the prey alone, then each agent of the team receives a penalty of -0.1 (each agent gets this penalty that will be added into a shared team reward). The results are smoothed with a 10-step cumulative moving average (Figure \ref{fig:pp4_pen}). 

In Figure \ref{fig:pp4_pen} we can see that RQN can still maintain a good performance over time, while the benchmarked methods suffer from this change on the environment, mainly QMIX. The results suggest that QMIX is more sensitive to monotonicity changes in the environment as it was proved before in \cite{son_qtran_2019}. RQN however is capable of still performing well even with a change in the level of monotonicty of the environment, suggesting that RQN eliminates this constraint imposed by QMIX. This observation is also supported by some of the main experiments in section \ref{sec:res} with some non-monotonic tasks. The hyperparameters used in this additional experimental setting are summarized in Table \ref{table2}.

\section{Conclusions and Future Work}\label{sec:conc}
We study the problem of value-function factorization for MARL within the scope of centralized training and decentralized execution scheme \cite{foerster_counterfactual_2017, lowe_multi-agent_2017}. Here, we propose a method that uses a novel agent configuration, named as Residual Q-Networks (RQNs), which transforms the individual factor functions in a way that preserves the Individual-Global-Max (IGM) property, and at the same time enables the factorization in a robust way. The RQN is designed as an auxiliary network that learns to adjust the action-value function trajectories of individual agents to reflect the relative importance of different episodes. With the proposed RQN based approach, it is possible to factorize a wider family of multi-agent environments by improving the credit assignment problem \cite{wei_multiagent_2018,yang_cm3_2020}, an issue that is prominent in decentralized learning. The proposed RQN architecture not only achieves better performance, but also improves the performance stability over time. 

The experimental results illustrate that the proposed method can outperform the state-of-the-art methods such as QTRAN, VDN, QMIX, QPLEX and WQMIX in certain tasks with strong punishments for non-cooperative behaviours and with varying levels of complexity and with. An exception to this is in StarCraft environments, where complete state information is available, but RQN is yet competitive and learns to solve the task. The results suggest that the proposed RQN based approach is capable of factorizing value-functions for more families of environments than the previous method, owing to a more robust and relaxed transformation that preserves IGM, and a neural architecture that is more tractable. Moreover, opposing to most of the benchmarked methods, RQN does not take advantage of full state information during the training phase. This enhances the practical application of RQN, especially towards non-simulated environments where complete state information is unavailable even during training time.
Future work involves the scalability of MARL towards more complex environments, where the number of agents gradually increases. Accordingly, we aim to investigate how curriculum-based approaches can be used to extend the current method.   

\appendix[Win rates for the Starcraft II environments]\label{appendix}
\begin{figure}[!htb]\label{win_rates}
    \centering
    \includegraphics[width=\columnwidth]{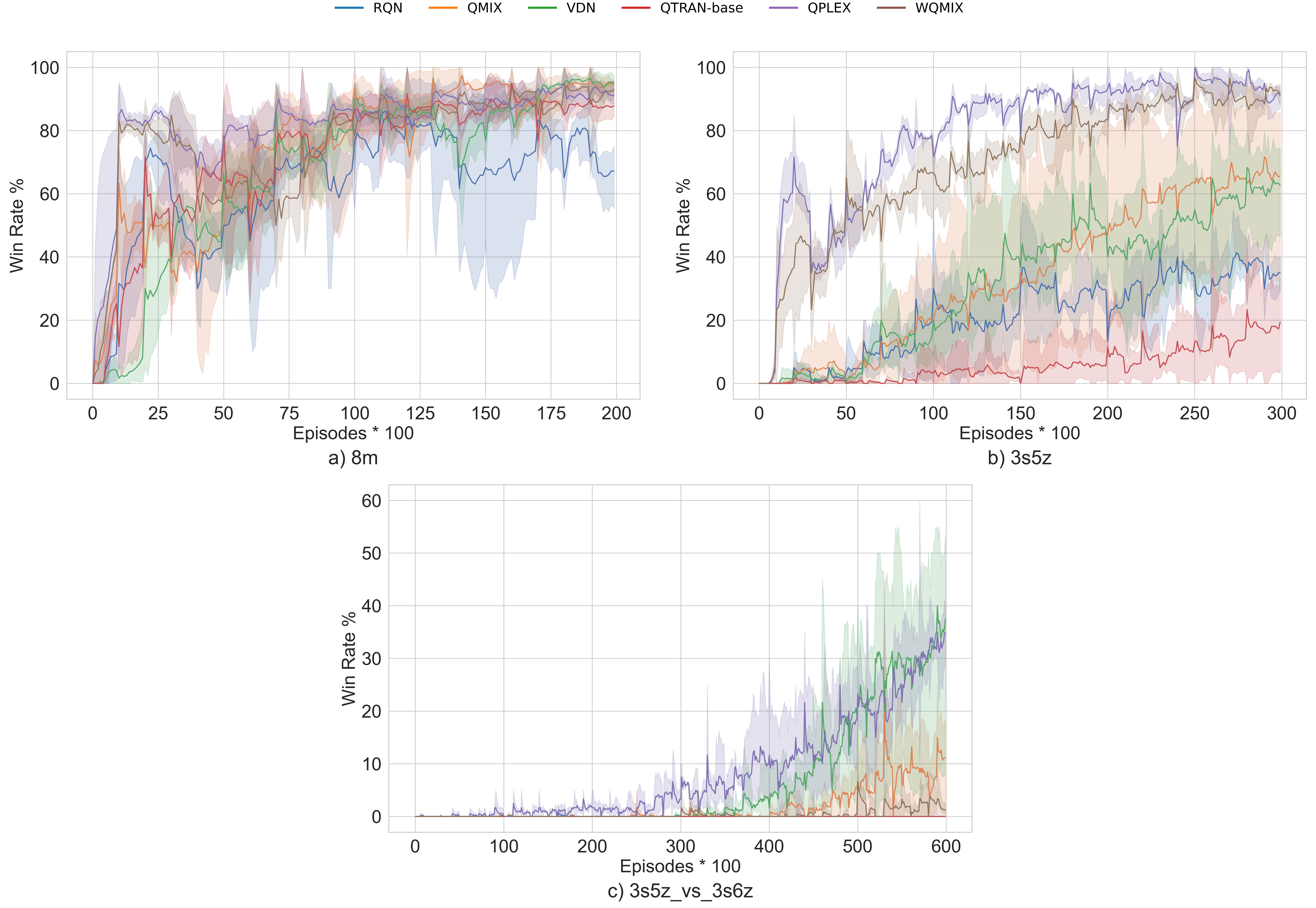}
    \caption{Win rates for the experimented methods in the Starcraft II environments. The bold area represents the 95\% confidence intervals.}
    \label{fig:smac_win_rates}
\end{figure}

\vskip -2\baselineskip plus -1fil
\begin{IEEEbiographynophoto}{Rafael Pina}
is currently a Ph.D. student at Loughborough University London in machine intelligence. His work focus mainly on Multi-Agent Reinforcement Learning. He got his B.S. in Informatics Engineering at the University of Coimbra in 2019 and his M.S. in Cyber Security and Big Data at Loughborough University London in 2020.
\end{IEEEbiographynophoto}
\vskip -2\baselineskip plus -1fil
\begin{IEEEbiographynophoto}{Varuna De Silva}
received his Ph.D. from University of Surrey in United
Kingdom in 2011. He was a postdoctoral researcher in the same institute between 2011-2013, working on video and image processing for Multiview video broadcasting applications. Since 2013 November, he worked as a senior algorithms developer for image signal processors at Apical Ltd, UK (Now part of ARM Plc). In April 2016, he joined Loughborough University, as a lecturer in digital engineering, where he was promoted to a Senior Lecturer in Jan 2020. 
\end{IEEEbiographynophoto}
\vskip -2\baselineskip plus -1fil
\begin{IEEEbiographynophoto}{Joosep Hook} holds a B.S. in Computer Engineering from the University of Tartu. He is currently pursuing his Ph.D. in machine learning at Loughborough University London. His work focus on Multi-Agent Reinforcement Learning and Curriculum Learning.  
\end{IEEEbiographynophoto}
\vskip -2\baselineskip plus -1fil
\begin{IEEEbiographynophoto}{Ahmet Kondoz} received the B.S. (Hons.) degree in engineering from the University of
Greenwich, Greenwich, U.K., in 1983, and the Ph.D. degree in telecommunication from the University of Surrey, Guildford, U.K., in 1987. He became a Lecturer in 1988, a Reader in 1995, and then a Professor in Multimedia Communication Systems in 1996, at the University of Surrey. Ahmet took part in setting up of the world renowned Centres CCSR and the I-Lab at the University of Surrey before joining Loughborough University London where he is now the Director of the Institute for Digital Technologies. He has published more than 400 journal and conference papers, three books, and nine patents.
\end{IEEEbiographynophoto}







\end{document}